\documentclass[runningheads]{llncs}

\input{psfig.sty}

\begin{document}

\pagestyle{headings}

\mainmatter

\title{Speech-Driven Text Retrieval: Using Target IR
Collections for Statistical Language Model Adaptation in Speech
Recognition}

\titlerunning{Speech-Driven Text Retrieval}

\author{Atsushi Fujii\inst{1} \and Katunobu Itou\inst{2}
\and Tetsuya Ishikawa\inst{1}}

\authorrunning{Atsushi Fujii et al.}

\institute{University of Library and Information Science\\
1-2 Kasuga, Tsukuba, 305-8550, Japan\\
\email{\{fujii,ishikawa\}@ulis.ac.jp}\\
\and
National Institute of Advanced Industrial Science and Technology\\
1-1-1 Chuuou Daini Umezono, Tsukuba, 305-8568, Japan\\
\email{itou@ni.aist.go.jp}}

\maketitle

\begin{abstract}
  Speech recognition has of late become a practical technology for
  real world applications. Aiming at speech-driven text retrieval,
  which facilitates retrieving information with spoken queries, we
  propose a method to integrate speech recognition and retrieval
  methods. Since users speak contents related to a target collection,
  we adapt statistical language models used for speech recognition
  based on the target collection, so as to improve both the
  recognition and retrieval accuracy. Experiments using existing test
  collections combined with dictated queries showed the effectiveness
  of our method.
\end{abstract}

\newcommand{\etal}{et~al.}
\newcommand{\etaleos}{et~al}
\newcommand{\eq}[1]{(\ref{#1})}

\section{Introduction}
\label{sec:introduction}

Automatic speech recognition, which decodes human voice to generate
transcriptions, has of late become a practical technology.  It is
feasible that speech recognition is used in real world computer-based
applications, specifically, those associated with human language.  In
fact, a number of speech-based methods have been explored in the
information retrieval community, which can be classified into the
following two fundamental categories:
\begin{itemize}
\item spoken document retrieval, in which written queries are used to
  search speech (e.g., broadcast news audio) archives for relevant
  speech information~\cite{johnson:icassp-99,jones:sigir-96,sheridan:sigir-97,singhal:sigir-99,srinivasan:sigir-2000,wechsler:sigir-98,whittaker:sigir-99},
\item speech-driven (spoken query) retrieval, in which spoken queries
  are used to retrieve relevant textual information~\cite{barnett:eurospeech-97,crestani:fqas-2000}.
\end{itemize}

Initiated partially by the TREC-6 spoken document retrieval (SDR)
track~\cite{garofolo:trec-97}, various methods have been proposed for
spoken document retrieval.  However, a relatively small number of
methods have been explored for speech-driven text retrieval, although
they are associated with numerous keyboard-less retrieval
applications, such as telephone-based retrieval, car navigation
systems, and user-friendly interfaces.

Barnett~\etal~\cite{barnett:eurospeech-97} performed comparative
experiments related to speech-driven retrieval, where an existing
speech recognition system was used as an input interface for the
INQUERY text retrieval system.  They used as test inputs 35 queries
collected from the TREC 101-135 topics, dictated by a single male
speaker.  Crestani~\cite{crestani:fqas-2000} also used the above 35
queries and showed that conventional relevance feedback techniques
marginally improved the accuracy for speech-driven text retrieval.

These above cases focused solely on improving text retrieval methods
and did not address problems of improving speech recognition accuracy.
In fact, an existing speech recognition system was used with no
enhancement. In other words, speech recognition and text retrieval
modules were fundamentally independent and were simply connected by
way of an input/output protocol.

However, since most speech recognition systems are trained based on
specific domains, the accuracy of speech recognition across domains is
not satisfactory. Thus, as can easily be predicted, in cases of
Barnett~\etal~\cite{barnett:eurospeech-97} and
Crestani~\cite{crestani:fqas-2000}, a relatively high speech
recognition error rate considerably decreased the retrieval accuracy.
Additionally, speech recognition with a high accuracy is crucial for
interactive retrieval.

Motivated by these problems, in this paper we integrate (not simply
connect) speech recognition and text retrieval to improve both
recognition and retrieval accuracy in the context of speech-driven
text retrieval.

Unlike general-purpose speech recognition aimed to decode any
spontaneous speech, in the case of speech-driven text retrieval, users
usually speak contents associated with a target collection, from which
documents relevant to their information need are retrieved.  In a
stochastic speech recognition framework, the accuracy depends
primarily on acoustic and language models~\cite{bahl:ieee-tpami-1983}.
While acoustic models are related to phonetic properties, language
models, which represent linguistic contents to be spoken, are strongly
related to target collections.  Thus, it is intuitively feasible that
language models have to be produced based on target collections.

To sum up, our belief is that by adapting a language model based on a
target IR collection, we can improve the speech recognition and text
retrieval accuracy, simultaneously.

Section~\ref{sec:system} describes our prototype speech-driven text
retrieval system, which is currently implemented for Japanese.
Section~\ref{sec:experimentation} elaborates on comparative
experiments, in which existing test collections for Japanese text
retrieval are used to evaluate the effectiveness of our system.

\section{System Description}
\label{sec:system}

\subsection{Overview}
\label{subsec:system_overview}

\begin{figure}[htbp]
  \begin{center}
    \leavevmode \psfig{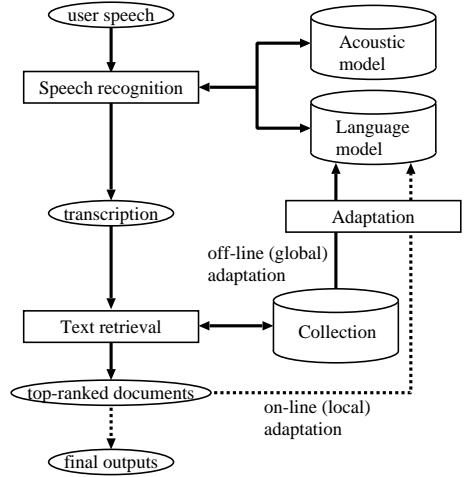}
  \end{center}
  \caption{The overall design of our speech-driven text retrieval system.}
  \label{fig:system}
\end{figure}

Figure~\ref{fig:system} depicts the overall design of our
speech-driven text retrieval system, which consists of speech
recognition, text retrieval and adaptation modules. We explain the
retrieval process based on this figure.

In the off-line process, the adaptation module uses the entire target
collection (from which relevant documents are retrieved) to produce a
language model, so that user speech related to the collection can be
recognized with a high accuracy.  On the other hand, an acoustic model
is produced independent of the target collection.

In the on-line process, given an information need spoken by a user,
the speech recognition module uses the acoustic and language models to
generate a transcription for the user speech.  Then, the text
retrieval module searches the collection for documents relevant to the
transcription, and outputs a specific number of top-ranked documents
according to the degree of relevance, in descending order.

These documents are fundamentally final outputs. However, in the case
where the target collection consists of multiple domains, a language
model produced in the off-line adaptation process is not necessarily
precisely adapted to a specific information need.  Thus, we optionally
use top-ranked documents obtained in the initial retrieval process for
an on-line adaptation, because these documents are associated with the
user speech more than the entire collection.  We then re-perform
speech recognition and text retrieval processes to obtain final
outputs.

In other words, our system is based on the two-stage retrieval
principle~\cite{kwok:sigir-98}, where top-ranked documents retrieved
in the first stage are intermediate results, and are used to improve
the accuracy for the second (final) stage.  From a different
perspective, while the off-line adaptation process produces the {\it
global\/} language model for a target collection, the on-line
adaptation process produces a {\it local\/} language model based on
the user speech.

In the following sections, we explain speech recognition, adaptation,
and text retrieval modules in
Figure~\ref{fig:system}, respectively.

\subsection{Speech Recognition}
\label{subsec:speech_recognition}

The speech recognition module generates word sequence $W$, given
phoneme sequence $X$.  In the stochastic speech recognition framework,
the task is to output the $W$ maximizing $P(W|X)$, which is
transformed as in equation~\eq{eq:bayes} through use of the Bayesian
theorem.
\begin{equation}
  \label{eq:bayes}
  \arg\max_{W}P(W|X) = \arg\max_{W}P(X|W)\cdot P(W)
\end{equation}
Here, $P(X|W)$ models a probability that word sequence $W$ is
transformed into phoneme sequence $X$, and $P(W)$ models a probability
that $W$ is linguistically acceptable. These factors are usually
called acoustic and language models, respectively.

For the speech recognition module, we use the Japanese dictation
toolkit~\cite{kawahara:icslp-2000}\footnote{http://winnie.kuis.kyoto-u.ac.jp/dictation/},
which includes the ``Julius'' recognition engine and acoustic/language
models trained based on newspaper articles. This toolkit also includes
development softwares, so that acoustic and language models can be
produced and replaced depending on the application.  While we use the
acoustic model provided in the toolkit, we use new language models
produced by way of the adaptation process (see
Section~\ref{subsec:lm_adaptation}).

\subsection{Language Model Adaptation}
\label{subsec:lm_adaptation}

The basis of the adaptation module is to produce a word-based $N$-gram
(in our case, a combination of bigram and trigram) model by way of
source documents.

In the off-line (global) adaptation process, we use the ChaSen
morphological analyzer~\cite{matsumoto:chasen-99} to extract words
from the entire target collection, and produce the global $N$-gram
model.

On the other hand, in the on-line (local) adaptation process, only
top-ranked documents retrieved in the first stage are used as source
documents, from which word-based $N$-grams are extracted as performed
in the off-line process.  However, unlike the case of the off-line
process, we do not produce the entire language model. Instead, we
re-estimate only statistics associated with top-ranked documents, for
which we use the MAP (Maximum A-posteriori Probability) estimation
method~\cite{masataki:icassp-97}.

Although the on-line adaptation theoretically improves the retrieval
accuracy, for real-time usage, the trade-off between the retrieval
accuracy and computational time required for the on-line process has
to be considered.

Our method is similar to the one proposed by Seymore and
Rosenfeld~\cite{seymore:eurospeech-97} in the sense that both methods
adapt language models based on a small number of documents related to
a specific domain (or topic). However, unlike their method, our method
does not require corpora manually annotated with topic tags.

\subsection{Text Retrieval}
\label{subsec:text_retrieval}

The text retrieval module is based on an existing probabilistic
retrieval method~\cite{robertson:sigir-94}, which computes the
relevance score between the transcribed query and each document in the
collection.  The relevance score for document $i$ is computed based on
equation~\eq{eq:okapi}.
\begin{equation}
  \label{eq:okapi}
  \sum_{t} \left(\frac{\textstyle TF_{t,i}}{\textstyle
    \frac{\textstyle DL_{i}}{\textstyle avglen} +
    TF_{t,i}}\cdot\log\frac{\textstyle N}{\textstyle DF_{t}}\right)
\end{equation}
Here, $t$'s denote terms in transcribed queries.  $TF_{t,i}$ denotes
the frequency that term $t$ appears in document $i$. $DF_{t}$ and $N$
denote the number of documents containing term $t$ and the total
number of documents in the collection. $DL_{i}$ denotes the length of
document $i$ (i.e., the number of characters contained in $i$), and
$avglen$ denotes the average length of documents in the collection.

We use content words extracted from documents as terms, and perform a
word-based indexing. For this purpose, we use the ChaSen morphological
analyzer~\cite{matsumoto:chasen-99} to extract content words. We
extract terms from transcribed queries using the same method.

\section{Experimentation}
\label{sec:experimentation}

\subsection{Test Collections}
\label{subsec:test_collection}

We investigated the performance of our system based on the NTCIR
workshop evaluation methodology, which resembles the one in the TREC
ad hoc retrieval track. In other words, each system outputs 1,000 top
documents, and the TREC evaluation software was used to plot
recall-precision curves and calculate non-interpolated average
precision values.

The NTCIR workshop was held twice (in 1999 and 2001), for which two
different test collections were produced: the NTCIR-1 and 2
collections~\cite{ntcir-99,ntcir-2001}\footnote{http://research.nii.ac.jp/\~{}ntcadm/index-en.html}.
However, since these collections do not include spoken queries, we
asked four speakers (two males/females) to dictate information needs
in the NTCIR collections, and simulated speech-driven text retrieval.

The NTCIR collections include documents collected from technical
papers published by 65 Japanese associations for various fields. Each
document consists of the document ID, title, name(s) of author(s),
name/date of conference, hosting organization, abstract and author
keywords, from which we used titles, abstracts and keywords for the
indexing. The number of documents in the NTCIR-1 and 2 collections are
332,918 and 736,166, respectively (the NTCIR-1 documents are a subset
of the NTCIR-2).

The NTCIR-1 and 2 collections also include 53 and 49 topics,
respectively. Each topic consists of the topic ID, title of the topic,
description, narrative.  Figure~\ref{fig:topic} shows an English
translation for a fragment of the NTCIR topics\footnote{The NTCIR-2
collection contains Japanese topics and their English translations.},
where each field is tagged in an SGML form. In general, titles are not
informative for the retrieval. On the other hand, narratives, which
usually consist of several sentences, are too long to speak. Thus,
only descriptions, which consist of a single phrase and sentence, were
dictated by each speaker, so as to produce four different sets of 102
spoken queries.

\begin{figure*}[htbp]
  \begin{center}
    \leavevmode
    \small
    \begin{quote}
      \tt
      <TOPIC q=0118>\\
      <TITLE>TV conferencing</TITLE>\\
      <DESCRIPTION>Distance education support systems using TV
      conferencing</DESCRIPTION>\\
      <NARRATIVE>A relevant document will provide information on the
      development of distance education support systems using TV
      conferencing. Preferred documents would present examples of
      using TV conferencing and discuss the results. Any reported
      methods of aiding remote teaching are relevant documents (for
      example, ways of utilizing satellite communication, the
      Internet, and ISDN circuits).</NARRATIVE>\\
      </TOPIC>
    \end{quote}
    \caption{An English translation for an example topic in the NTCIR
      collections.}
    \label{fig:topic}
  \end{center}
\end{figure*}

In the NTCIR collections, relevance assessment was performed based on
the pooling method~\cite{voorhees:sigir-98}. First, candidates for
relevant documents were obtained with multiple retrieval
systems. Then, for each candidate document, human expert(s) assigned
one of three ranks of relevance: ``relevant,'' ``partially relevant''
and \mbox{``irrelevant.''} The NTCIR-2 collection also includes
``highly relevant'' documents. In our evaluation, ``highly relevant''
and ``relevant'' documents were regarded as relevant ones.

\subsection{Comparative Evaluation}
\label{subsec:comparison}

In order to investigate the effectiveness of the off-line language
model adaptation, we compared the performance of the following
different retrieval methods:
\begin{itemize}
\item text-to-text retrieval, which used written descriptions
  as queries, and can be seen as the perfect speech-driven text retrieval,
\item speech-driven text retrieval, in which a language model produced
  based on the NTCIR-2 collection was used,
\item speech-driven text retrieval, in which a language model produced
  based on ten years worth of {\it Mainichi Shimbun\/} Japanese newspaper
  articles (1991-2000) was used.
\end{itemize}
The only difference in producing two different language models (i.e.,
those based on the NTCIR-2 collection and newspaper articles) are the
source documents. In other words, both language models have the same
vocabulary size (20,000), and were produced using the same softwares.

Table~\ref{tab:lang_model} shows statistics related to word
tokens/types in two different source corpora for language modeling,
where the line ``Coverage'' denotes the ratio of word tokens contained
in the resultant language model. Most of word tokens were covered in
both language models.

\begin{table}[htbp]
  \begin{center}
    \caption{Statistics associated with source words for language
    modeling.}
    \medskip
    \leavevmode
    \small
    \tabcolsep=3pt
    \begin{tabular}{lcc} \hline\hline
      & NTCIR & News \\ \hline
      \# of Types & 454K & 315K \\
      \# of Tokens & 175M & 262M \\
      Coverage & 97.9\% & 96.5\% \\
      \hline
    \end{tabular}
    \label{tab:lang_model}
  \end{center}
\end{table}

In cases of speech-driven text retrieval methods, queries dictated by
four speakers were used individually. Thus, in practice we compared
nine different retrieval methods. Although the Julius decoder outputs
more than one transcription candidate for a single speech input, we
used only the one with the greatest probability score. The results did
not significantly change depending on whether or not we used
lower-ranked transcriptions as queries.

Table~\ref{tab:results} shows the non-interpolated average precision
values and word error rate in speech recognition, for different
retrieval methods. As with existing experiments for speech
recognition, word error rate (WER) is the ratio between the number of
word errors (i.e., deletion, insertion, and substitution) and the
total number of words. In addition, we also investigated error rate
with respect to query terms (i.e., keywords used for retrieval), which
we shall call ``term error rate (TER).''

In Table~\ref{tab:results}, the first line denotes results of the
text-to-text retrieval, which were relatively high compared with
existing results reported in the NTCIR
workshops~\cite{ntcir-99,ntcir-2001}.

\begin{table*}[htbp]
  \begin{center}
    \caption{Results for different retrieval methods (AP: average
    precision, WER: word error rate, TER: term error rate).}
    \medskip
    \leavevmode
    \small
    \tabcolsep=5pt
    \begin{tabular}{lcccccc} \hline\hline
      & \multicolumn{3}{c}{NTCIR-1} & \multicolumn{3}{c}{NTCIR-2} \\
      \cline{2-7}
      {\hfill\centering Method\hfill}
      & AP & WER & TER
      & AP & WER & TER \\ \hline
      Text & 0.3320 & --- & --- & 0.3118 & --- & --- \\
      M1 (NTCIR) & 0.2708 & 0.1659 & 0.2190 & 0.2504 & 0.1532 & 0.2313 \\
      M2 (NTCIR) & 0.2471 & 0.2034 & 0.2381 & 0.2114 & 0.2180 & 0.2799 \\
      F1 (NTCIR) & 0.2276 & 0.1961 & 0.2857 & 0.1873 & 0.1885 & 0.2500 \\
      F2 (NTCIR) & 0.2642 & 0.1477 & 0.2222 & 0.2376 & 0.1635 & 0.2388 \\
      M1 (News) &  0.1076 & 0.3547 & 0.5143 & 0.0790 & 0.3594 & 0.5149 \\
      M2 (News) &  0.1257 & 0.4044 & 0.5460 & 0.0691 & 0.5022 & 0.6343 \\
      F1 (News) &  0.1156 & 0.3801 & 0.5238 & 0.0798 & 0.4418 & 0.5709 \\
      F2 (News) &  0.1225 & 0.3317 & 0.5016 & 0.0917 & 0.4080 & 0.5858 \\
      \hline
    \end{tabular}
    \label{tab:results}
  \end{center}
\end{table*}

The remaining lines denote results of speech-driven text retrieval
combined with the NTCIR-based language model (lines 2-5) and the
newspaper-based model (lines 6-9), respectively.  Here, ``Mx'' and
``Fx'' denote male/female speakers, respectively. Suggestions which
can be derived from these results are as follows.

First, for both language models, results did not significantly change
depending on the speaker. The best average precision values for
speech-driven text retrieval were obtained with a combination of
queries dictated by a male speaker (M1) and the NTCIR-based language
model, which were approximately 80\% of those with the text-to-text
retrieval.

Second, by comparing results of different language models for each
speaker, one can see that the NTCIR-based model significantly
decreased WER and TER obtained with the newspaper-based model, and
that the retrieval method using the NTCIR-based model significantly
outperformed one using the newspaper-based model. In addition, these
results were observable, irrespective of the speaker.  Thus, we
conclude that adapting language models based on target collections was
quite effective for speech-driven text retrieval.

Third, TER was generally higher than WER irrespective of the speaker.
In other words, speech recognition for content words was more
difficult than functional words, which were not contained in query
terms.

We analyzed transcriptions for dictated queries, and found that speech
recognition error was mainly caused by the out-of-vocabulary
problem. In the case where major query terms are mistakenly
recognized, the retrieval accuracy substantially decreases.  In
addition, descriptions in the NTCIR topics often contain expressions
which do not appear in the documents, such as ``I want papers
about...''  Although these expressions usually do not affect the
retrieval accuracy, misrecognized words affect the recognition
accuracy for remaining words including major query
terms. Consequently, the retrieval accuracy decreases due to the
partial misrecognition.

\begin{figure}[htbp]
  \begin{center}
    \leavevmode \psfig{file=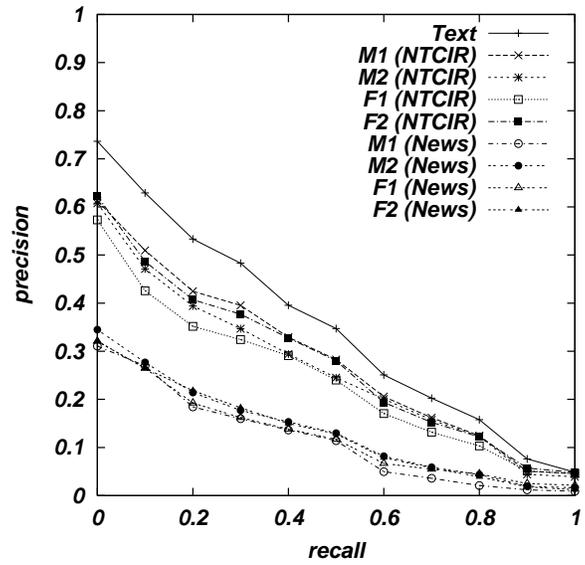,height=3in}
  \end{center}
  \caption{Recall-precision curves for different retrieval methods
  using the NTCIR-1 collection.}
  \label{fig:ntcir1}
\end{figure}

\begin{figure}[htbp]
  \begin{center}
    \leavevmode \psfig{file=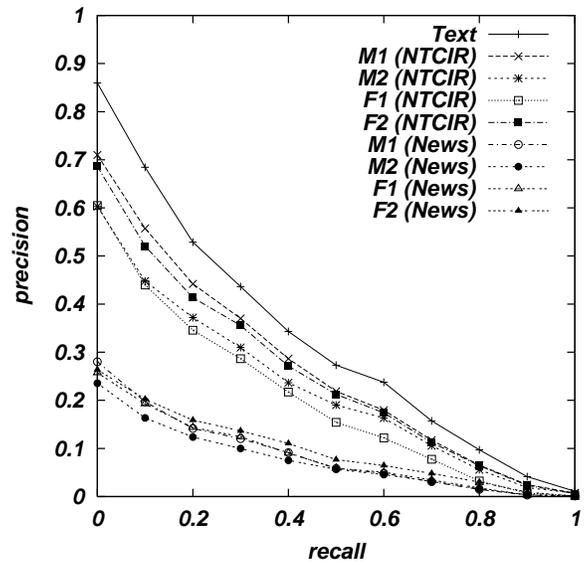,height=3in}
  \end{center}
  \caption{Recall-precision curves for different retrieval methods
  using the NTCIR-2 collection.}
  \label{fig:ntcir2}
\end{figure}

Finally, we investigated the trade-off between recall and precision.
Figures~\ref{fig:ntcir1} and \ref{fig:ntcir2} show recall-precision
curves of different retrieval methods, for the NTCIR-1 and 2
collections, respectively. In these figures, the relative superiority
for precision values due to different language models in
Table~\ref{tab:results} was also observable, regardless of the recall.

However, the effectiveness of the on-line adaptation remains an open
question and needs to be explored.

\section{Conclusion}
\label{sec:conclusion}

Aiming at speech-driven text retrieval with a high accuracy, we
proposed a method to integrate speech recognition and text retrieval
methods, in which target text collections are used to adapt
statistical language models for speech recognition.  We also showed
the effectiveness of our method by way of experiments, where dictated
information needs in the NTCIR collections were used as queries to
retrieve technical abstracts.  Future work would include experiments
on various collections, such as newspaper articles and Web pages.

\section{Acknowledgments}

The authors would like to thank the National Institute of Informatics
for their support with the NTCIR collections.

\bibliographystyle{abbrv}

\end{document}